%% file: main.tex
\documentclass[final]{cvpr}

\usepackage{times}
\usepackage{epsfig}
\usepackage{graphicx}
\usepackage{amsmath}
\usepackage{amssymb}
\usepackage{subfigure}
\usepackage[belowskip=-6pt,aboveskip=5pt]{caption}
\usepackage{stackengine}
\usepackage{cuted}
\usepackage{capt-of}
\usepackage{enumitem}

\usepackage{multirow}
\usepackage[table,xcdraw]{xcolor}

\usepackage[pagebackref=true,breaklinks=true,colorlinks,bookmarks=false]{hyperref}

\def\cvprPaperID{644} %
\def\confYear{CVPR 2021}

\begin{document}

\title{Bridging the Visual Gap: Wide-Range Image Blending}

\author{Chia-Ni Lu$\quad\quad\quad$Ya-Chu Chang$\quad\quad\quad$Wei-Chen Chiu\\
National Chiao Tung University (NCTU), Taiwan\\
MediaTek-NCTU Research Center, Taiwan\\
{\tt\small julialu67.cs08g@nctu.edu.tw$\quad$jenna.cs07g@nctu.edu.tw $\quad$walon@cs.nctu.edu.tw}
}

\twocolumn[{
\maketitle
\begin{center}
    \centering
    \vspace{-1.5em}
\stackunder[5pt]{\includegraphics[width=0.251\columnwidth]{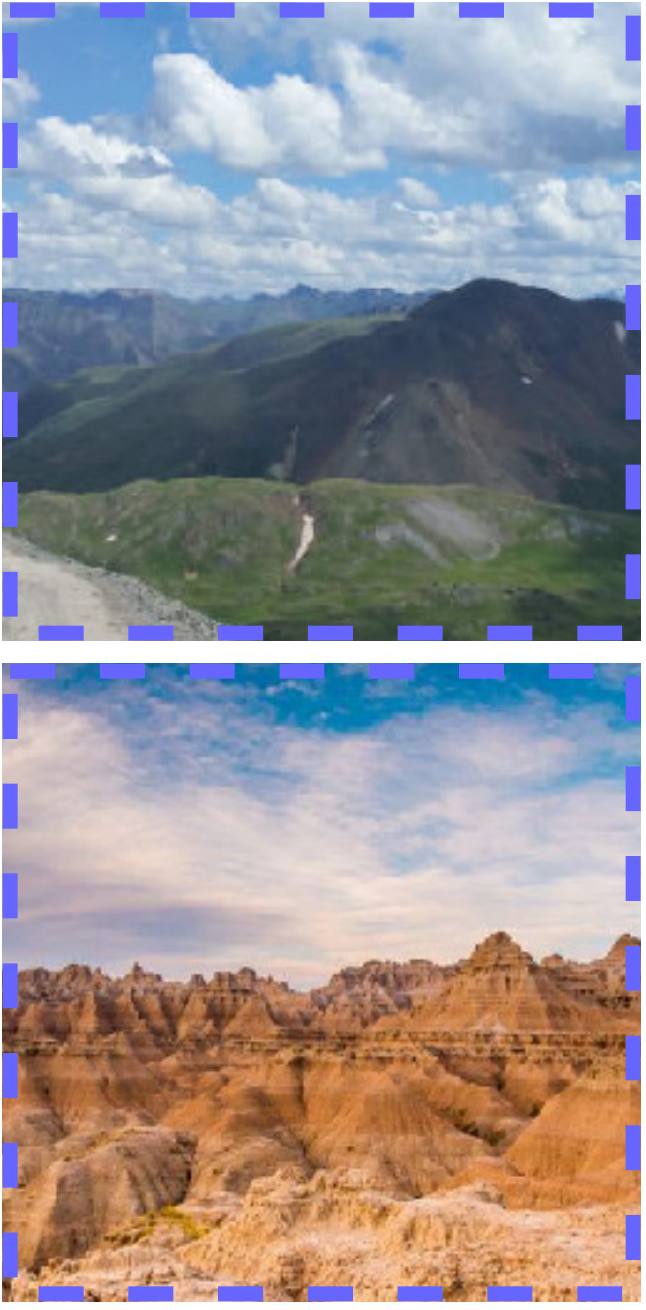}}{\footnotesize{First Input Photo}}
\hspace{-1mm}
\stackunder[5pt]{\includegraphics[width=0.251\columnwidth]{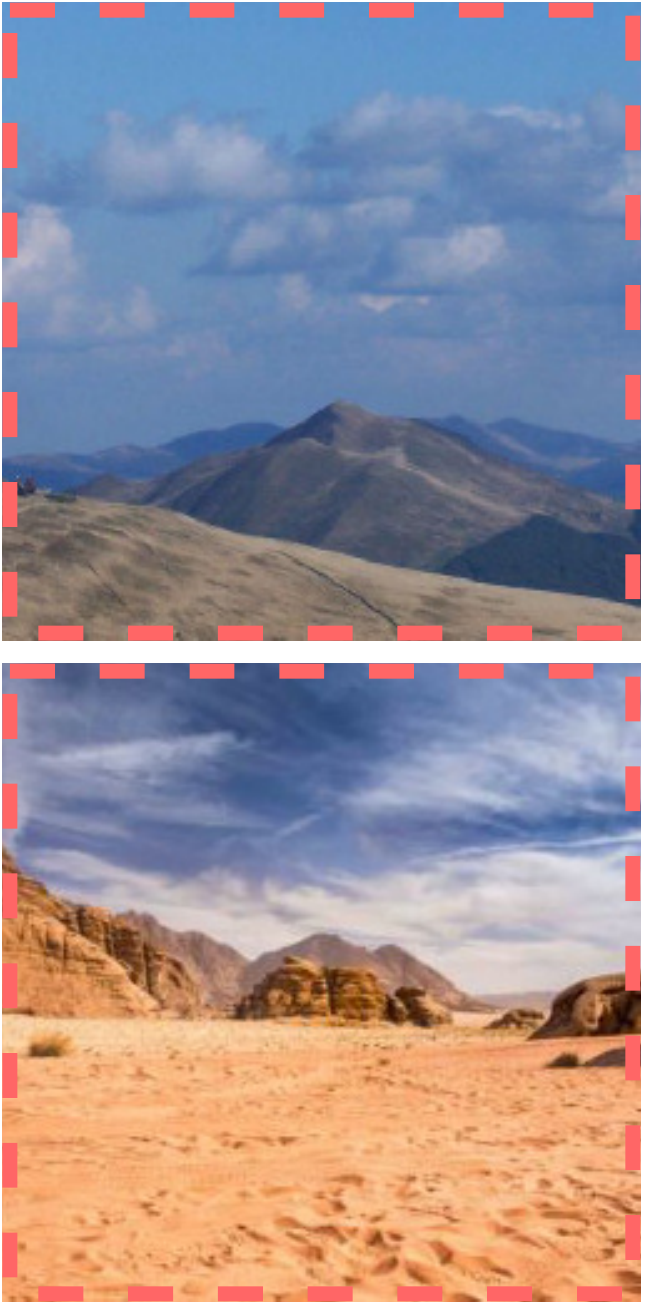}}{\footnotesize{Second Input Photo}}
\hspace{-1mm}
\stackunder[5pt]{\includegraphics[width=0.748\columnwidth]{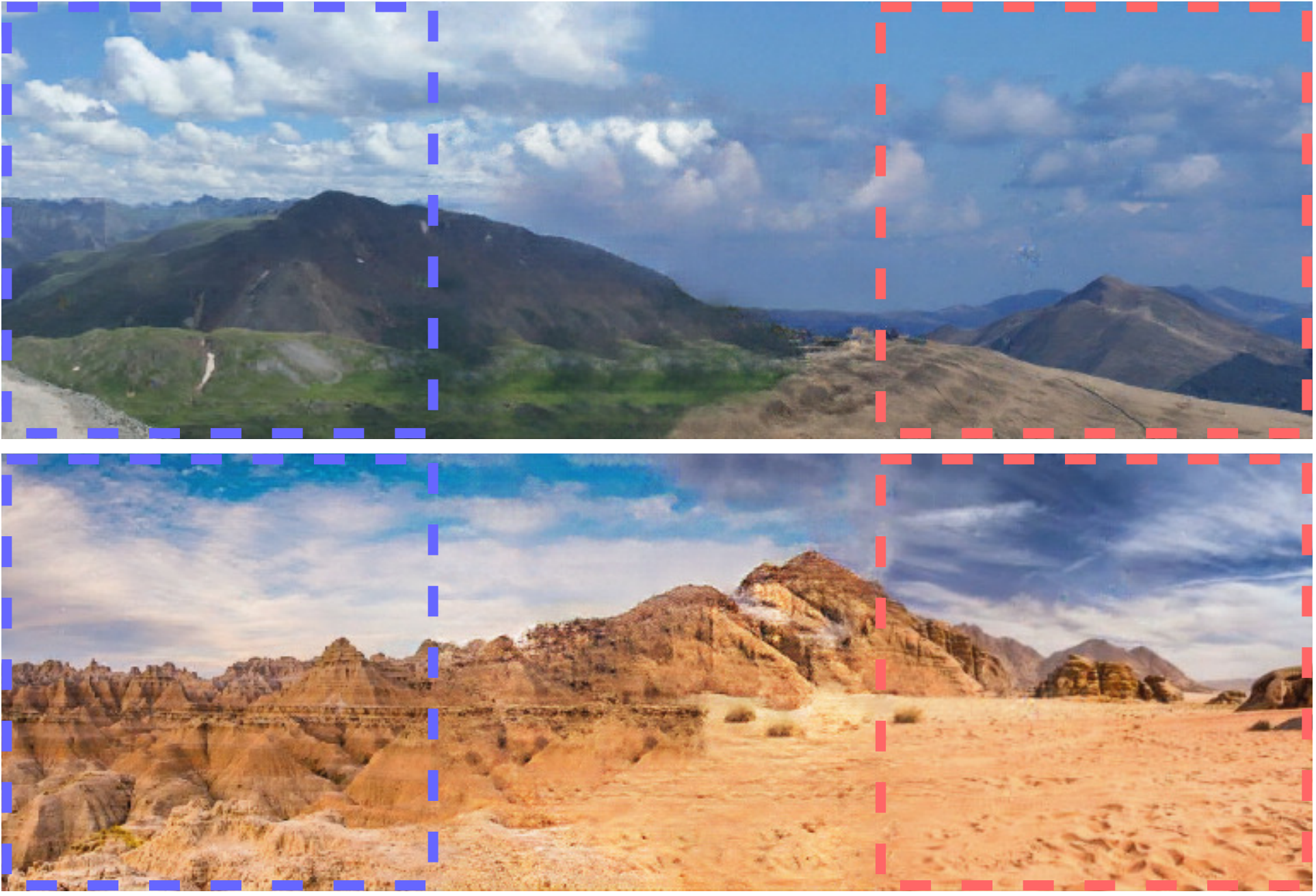}}{\footnotesize{Resultant Panorama-1}}
\hspace{-1mm}
\stackunder[5pt]{\includegraphics[width=0.748\columnwidth]{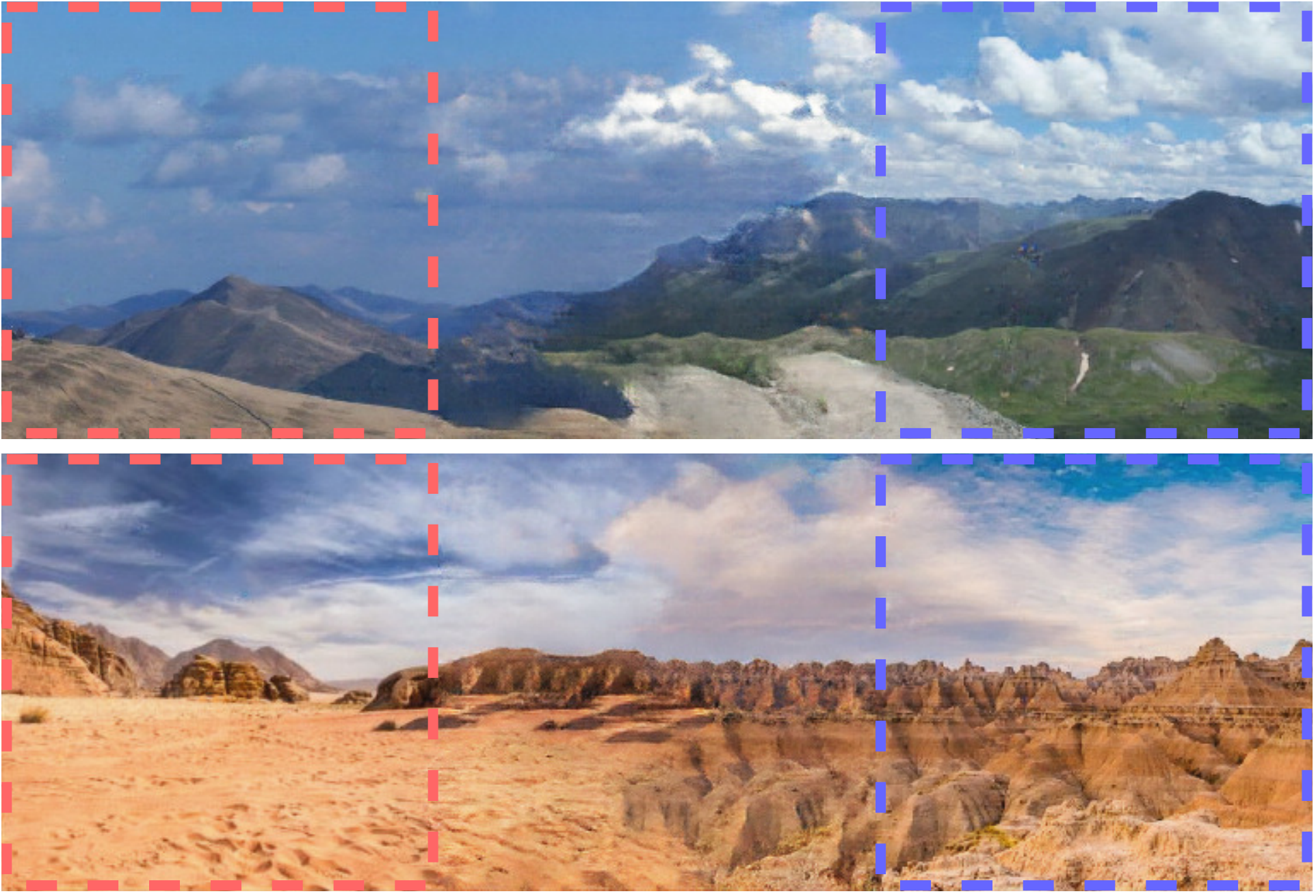}}{\footnotesize{Resultant Panorama-2}}
    \captionsetup{hypcap=false} 
    \captionof{figure}{Given two different input photos (the first and the second columns), our wide-range image blending model is able to seamlessly blend them into a novel panoramic image by generating smooth transition in the intermediate region between them. Here we show several examples of the resultant panoramas, where the result in the third column is produced by putting the first input photo (highlighted by blue dashed lines) on the left and the second input photo (highlighted by red dashed lines) on the right, while the fourth column is obtained with using opposite spatial arrangement.
    }
    \label{fig:teaser}
\end{center}
\vspace{.5em}
}]
\captionsetup{hypcap=true}

\begin{abstract}
\vspace{-.8em}
In this paper we propose a new problem scenario in image processing, wide-range image blending, which aims to smoothly merge two different input photos into a panorama by generating novel image content for the intermediate region between them. Although such problem is closely related to the topics of image inpainting, image outpainting, and image blending, none of the approaches from these topics is able to easily address it. We introduce an effective deep-learning model to realize wide-range image blending, where a novel Bidirectional Content Transfer module is proposed to perform the conditional prediction for the feature representation of the intermediate region via recurrent neural networks. In addition to ensuring the spatial and semantic consistency during the blending, we also adopt the contextual attention mechanism as well as the adversarial learning scheme in our proposed method for improving the visual quality of the resultant panorama. We experimentally demonstrate that our proposed method is not only able to produce visually appealing results for wide-range image blending, but also able to provide superior performance with respect to several baselines built upon the state-of-the-art image inpainting and outpainting approaches.

\end{abstract}

\input{1_intro.tex}

\input{2_related.tex}

\input{3_method.tex}

\input{4_experiment.tex}
\input{5_conclusion.tex}

{\small
\bibliographystyle{ieee_fullname}
\bibliography{egbib}
}

\end{document}

%% file: 1_intro.tex
\section{Introduction}
\label{sec:introduction}
Digital image processing, which carries out computer-based processing and manipulation on image data, has been playing an important role in our daily life, such as image inpainting for image restoration or object removal, image blending for image composition, and image outpainting (i.e.\ extrapolation) for digital content generation. In this paper, we propose a novel task of image processing: \textit{wide-range image blending}, in which it aims to smoothly merge two different images into a panorama by generating novel image content for the intermediate region between them, as shown in Figure~\ref{fig:teaser}. Such technique can contribute to bringing in more interesting ways for the content generation and image composition%
. For example, we could easily create a full panoramic image based on the photos taken by the front and rear cameras of a cellphone via applying wide-range image blending on them with two opposite spatial arrangements (i.e.\ one is putting the front photo on the left and the rear photo on the right, while the other one is opposite).

The main challenge of wide-range image blending lies in the requirement that the generated content for the intermediate region should be not only visually realistic but also semantically reasonable to achieve seamless transition from one input photo to another. Although there exists no approach for addressing wide-range image blending, such task is closely related to several topics of image processing. For instance, the extrapolation from input photos beyond their boundary towards the intermediate region fits exactly the scenario of image outpainting; by contrast, if we treat input photos as the given context and the intermediate region is what to be filled, the task of image inpainting appears.

However, no prior works of these topics is able to easily resolve wide-range image blending. For instance, although previous works for image inpainting~\cite{liu2018image, liu2019coherent, ren2019structureflow,wang2018image, yi2020contextual, yu2018generative, zeng2019learning, zeng2020high} are able to learn semantics from context and generate coherent structure for the missing region, they however could create artifacts
and blurry textures as the size of the missing region increases. Especially, if the content of two input photos is quite different, the inpainting approaches are also likely to have hard time on generating satisfactory results with smooth transition across input photos. On the other hand, even if we can apply the existing image outpainting model (e.g.~\cite{guo2020spiral, teterwak2019boundless, wang2019wide, yang2019very}) respectively on the two input photos for generating the image content of the intermediate region, there is no guarantee to have seamless composition between those two extrapolation results. Later in this paper, we will provide experimental evidence to demonstrate that directly adopting inpainting or outpainting methods without any modification leads to poor results under the problem scenario of wide-range image blending.

We propose a novel deep-learning-based model to perform wide-range image blending with all the aforementioned challenges/issues being well addressed. The architecture of our proposed model stems from the U-Net~\cite{ronneberger2015u} framework where the encoder takes two photos as input and the decoder outputs the resultant image of blending. Particularly, in the bottleneck of such U-Net-alike framework, we introduce a Bidirectional Content Transfer module for predicting the image content of the intermediate region, which is encouraged to ensure the continuity of the spatial configuration between the intermediate region and two input photos. Moreover, for making better use of the rich texture information from input photos and generating more delicate blending results, we propose to integrate the contextual attention mechanism~\cite{yu2018generative} on the skip connection between the encoder and the decoder. Last but not least,
we adopt adversarial learning~\cite{goodfellow2014generative} for improving the realness of the intermediate region, 
even when the two input photos are from significantly different scenes. It is also worth noting that our model learning does not require any supervision in the training data therefore being unsupervised. We conduct extensive ablation study to verify the contribution of our design choices, as well as provide both qualitative and quantitative comparisons with respect to several inpainting and outpainting baselines for demonstrating the efficacy of our proposed method in the task of wide-range image blending.

%% file: 2_related.tex
\section{Related Works}
\label{sec:related}

Image inpainting refers to filling missing regions of the corrupted input image and obtaining the visually realistic result. It has attracted much attention in the field of computer vision due to its wide applications. Numerous methods are proposed to address this task, for instance, \cite{zeng2019learning} proposes Pyramid-context Encoder Network, where a pyramid-context encoder learns the attention and fills the missing region from high-level semantic feature maps to low-level ones; \cite{ren2019structureflow} employs edge-preserved smooth images as additional
information to assist in the inpainting process.
In contrast to image inpainting, image outpainting aims to generate new content beyond the original boundaries for a given image. Previous works deal with this task from
different aspects. For example, 
\cite{wang2019wide} introduces a semantic regeneration network that learns semantic features from a small-size input and generates a full image; 
\cite{guo2020spiral} proposes SpiralNet which performs image outpainting in a spiral fashion, growing from an input sub-image along a spiral curve to an expanded full image. Although the above two research topics are highly correlated to our task of wide-range image blending, none of the existing approaches is able to generate intermediate region that bridges two different images with smooth transition and exquisite details.

%% file: 3_method.tex
\begin{figure*}[t]
	\centering
    \includegraphics[width=0.95\textwidth]{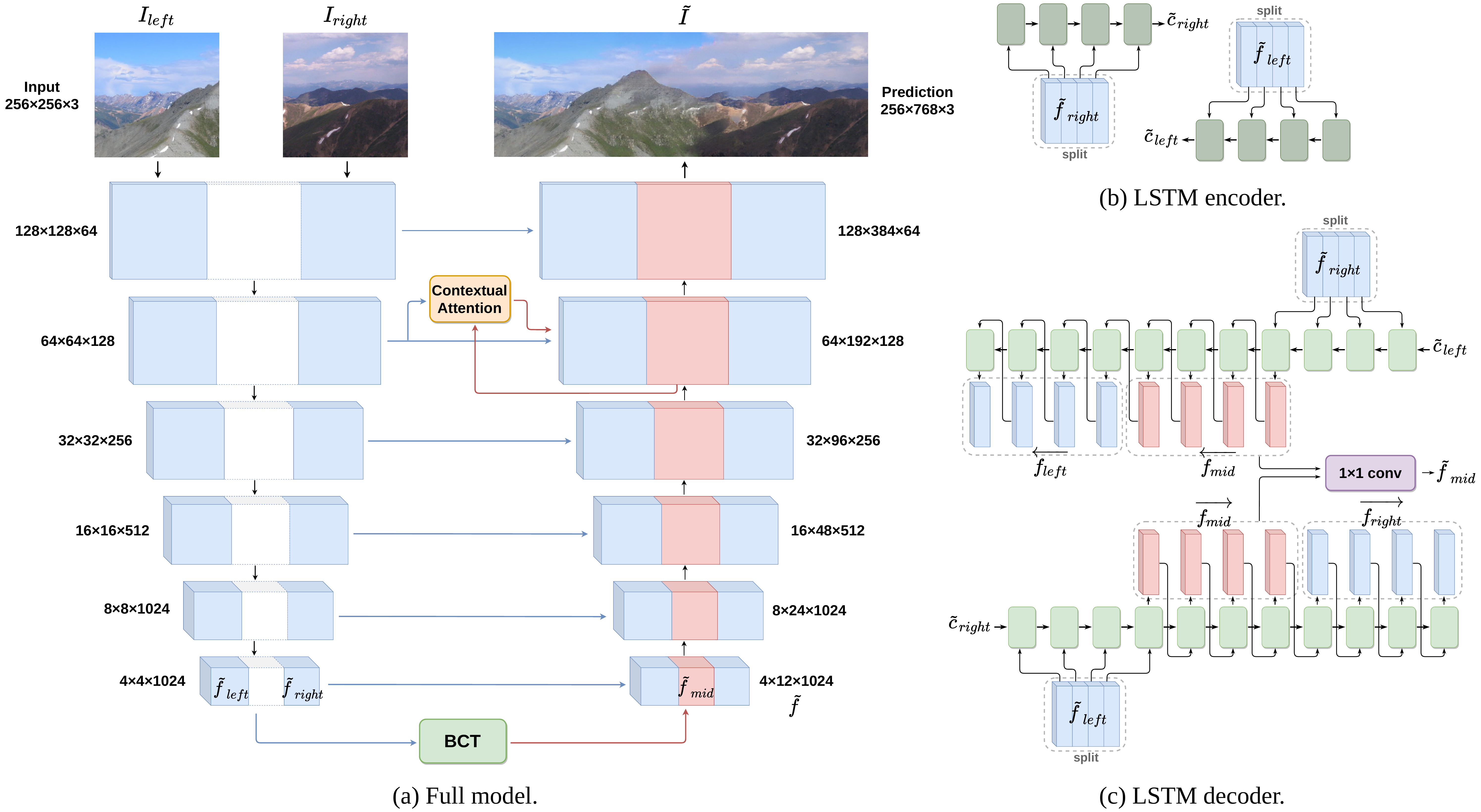}
    \caption{
    Illustration of our proposed model. (a) Our full model takes $I_{left}$ and $I_{right}$ as input, and compresses them into compact representations $\tilde{f}_{left}$ and $\tilde{f}_{right}$ individually via the encoder (cf. Section~\ref{encoder_decoder}). Afterwards, our novel Bidirectional Content Transfer (BCT) module is used to predict $\tilde{f}_{mid}$ from $\tilde{f}_{left}$ and $\tilde{f}_{right}$ (cf. Section~\ref{BCT} for more details). Lastly, based on the feature $\tilde{f}$, which is obtained by concatenating $\{\Tilde{f}_{left}, \tilde{f}_{mid}, \tilde{f}_{right}\}$ along the horizontal direction, the decoder generates our final result $\tilde{I}$. Noting that there is a contextual attention mechanism on the skip connection between the encoder and decoder, which helps to enrich the texture and details of our blending result (as described in Section~\ref{attention}). (b) The architecture of the LSTM encoder $\mathcal{E}_{BCT}$ in our BCT module, which encodes the information of $\tilde{f}_{left}$ or $\tilde{f}_{right}$ to generate $\tilde{c}_{left}$ or $\tilde{c}_{right}$. (c) The architecture of the conditional LSTM decoder $\mathcal{D}_{BCT}$ in our BCT module, which takes the condition $\tilde{c}_{right}$ (respectively $\tilde{c}_{left}$) as well as the input $\tilde{f}_{left}$ (respectively $\tilde{f}_{right}$) to predict the feature map $\protect\overrightarrow{f_{mid}}$ (respectively $\protect\overleftarrow{f_{mid}}$). The prediction of $\tilde{f}_{mid}$ related to the intermediate region, which blends between $\tilde{f}_{left}$ and $\tilde{f}_{right}$, is then obtained via concatenating $\protect\overrightarrow{f_{mid}}$ and $\protect\overleftarrow{f_{mid}}$ along the channel dimension followed by passing through a $1\times1$ convolutional layer.
    }
	\label{fig:model}
\end{figure*}

\section{Proposed Method}
\label{sec:method}
As motivated in previous sections, the objective of our proposed model of wide-range image blending is learning to generate new content for the intermediate region which connects two different input photos, thus leading to a semantically coherent and spatially smooth panoramic image. Our full model is shown in Figure~\ref{fig:model}, where in the following we will sequentially describe our model designs, including the image context encoder-decoder, the bidirectional content transfer module, and the contextual attention mechanism on skip connection, as well as the training details.

\subsection{Model Designs}
\label{model_design}
Given two input photos $I_{left}$ and $I_{right}$, our goal is to produce the wide-range image $\Tilde{I}$, which is obtained by horizontally concatenating three portions $\{\Tilde{I}_{left}, \tilde{I}_{mid}, \tilde{I}_{right}\}$ generated from our proposed model. Particularly, the resultant $\Tilde{I}_{left}$ and $\Tilde{I}_{right}$ should be identical to their corresponding ground truth $I_{left}$ and $I_{right}$ respectively, whereas $\tilde{I}_{mid}$ should provide smooth transition between $\tilde{I}_{left}$ and $\tilde{I}_{right}$. In order to generate the intermediate region $\tilde{I}_{mid}$ that is able to retain coherent spatial configuration with respect to the input photos while realizing the blending, yet still preserve the rich texture and details, we propose several designs to extract semantics and textures from the two input photos, and to incorporate those information into $\tilde{I}_{mid}$ thus achieving favourable output $\Tilde{I}$.

\vspace{-4mm}
\subsubsection{Image Context Encoder-Decoder}
\label{encoder_decoder}
\vspace{-2mm}
Our proposed model is U-Net-alike, in which we adopt the encoder-decoder part of a state-of-the-art network of image outpainting proposed by~\cite{yang2019very} as the basis for our model building. Basically, the network architectures of encoder and decoder are derived from ResNet-50~\cite{he2016deep}, with adding extra convolution and transpose-convolution layers in the encoder and decoder respectively to fit the size of our input photos, and removing all instance normalization to avoid the water-droplet-like artifacts~\cite{karras2020analyzing} (except for the ones in the three deepest convolution and transpose-convolution layers).
Also, there are skip connections for connecting the features between encoder and decoder at each layer (symmetric with respect to the bottleneck). Moreover, the techniques of Skip Horizontal Connection (SHC) and Global Residual Block (GRB), which are also from~\cite{yang2019very}, are exploited as well to improve the quality of output image, where SHC takes full advantage of the information extracted from the encoder and fuses it into the decoder, and GRB uses dilated convolutions to enlarge the receptive field in order to better strengthen the coherence among the input photos and the intermediate region along the network computation (please refer to the supplementary materials for detailed implementation of SHC, GRB and the encoder-decoder architecture in our proposed method). 

The encoder $\mathcal{E}$ extracts the feature representation of the input images, i.e.\ 
$\tilde{f}_{left} = \mathcal{E}(I_{left})$ and $\tilde{f}_{right} = \mathcal{E}(I_{right})$. After using the bidirectional content transfer module to predict the feature representation $\tilde{f}_{mid}$ related to the intermediate region $\tilde{I}_{mid}$, the decoder $\mathcal{D}$ takes $\Tilde{f}$ formed by horizontally concatenating $\{\tilde{f}_{left}, \tilde{f}_{mid}, \tilde{f}_{right}\}$ as input and generates the final wide-range image $\Tilde{I}$.

\vspace{-4mm}
\subsubsection{Bidirectional Content Transfer}
\label{BCT}
\vspace{-1mm}
The Bidirectional Content Transfer (BCT) module is a novel component proposed by us to predict $\tilde{f}_{mid}$ from $\tilde{f}_{left}$ and $\tilde{f}_{right}$. As the image content of the intermediate region should appear as a smooth transition from $I_{left}$ to $I_{right}$, we propose to first vertically and equally split $\tilde{f}_{left}$ into a sequence of sub-feature maps (i.e.\ with the same height and number of channels as $\tilde{f}_{left}$ but smaller width) then adopt the Long Short-Term Memory (LSTM) model to perform sequential prediction for generating $\tilde{f}_{mid}$. Moreover, owing to the fact that such generated $\tilde{f}_{mid}$ should be also smoothly connected to $\tilde{f}_{right}$ despite it is expanded from $\tilde{f}_{left}$, we propose to explicitly make the sequential prediction of LSTM being conditioned on $\tilde{f}_{right}$. Besides, since the procedure described above should also holds for the opposite direction (i.e.\ starting from $\tilde{f}_{right}$ to sequentially predict $\tilde{f}_{mid}$ while being conditioned on  $\tilde{f}_{left}$), our LSTM is designed to be bidirectional.

Our proposed Bidirectional Content Transfer module consists of a LSTM encoder $\mathcal{E}_{BCT}$ and a conditional LSTM decoder $\mathcal{D}_{BCT}$, as shown in Figure~\ref{fig:model}(b) and Figure~\ref{fig:model}(c) respectively.
We assume that all $\tilde{f}_{left}$, $\tilde{f}_{mid}$, and $\tilde{f}_{right}$ can be equally and vertically split into $K$ sub-feature maps, denoted as $\tilde{f}_{left} = \{f_{left}^k\}_{k=1}^K$, $\tilde{f}_{mid} = \{f_{mid}^k\}_{k=1}^K$, and $\tilde{f}_{right} = \{f_{right}^k\}_{k=1}^K$.

First, for the prediction along the direction from $\tilde{f}_{left}$ through $\tilde{f}_{mid}$ towards $\tilde{f}_{right}$, as it needs being conditioned on the information from $\tilde{f}_{right}$, we use the LSTM encoder $\mathcal{E}_{BCT}$ to sequentially aggregate $\{f_{right}^k\}_{k=1}^K$ into a latent code $\tilde{c}_{right}$. Then, with having $\tilde{c}_{right}$ as its initial condition, the conditional LSTM decoder $\mathcal{D}_{BCT}$ takes input $\{f_{left}^k\}_{k=1}^K$ and sequentially predicts $\overrightarrow{f_{mid}^k}$ and $\overrightarrow{f_{right}^k}$, where $k$ increases from $1$ to $K$ and the superscript $\rightarrow$ indicates the left-to-right direction. The procedure is written as:
\vspace{-1mm}
\begin{equation}
\label{equ:encode_BCT_left2right}
\begin{aligned}
\tilde{c}_{right} &= \mathcal{E}_{BCT}(\{f_{right}^k\}_{k=1}^K) \\
\left (\{\overrightarrow{f_{mid}^k}\}_{k=1}^K, \{\overrightarrow{f_{right}^k}\}_{k=1}^K \right ) &= \mathcal{D}_{BCT}(\{f_{left}^k\}_{k=1}^K, \tilde{c}_{right})%
\end{aligned}
\end{equation}

Second, for the prediction of the opposite direction from $\tilde{f}_{right}$ through $\tilde{f}_{mid}$ towards $\tilde{f}_{left}$, we perfor:
\vspace{-3mm}
\begin{equation}
\label{equ:encode_BCT_right2left}
\begin{aligned}
\tilde{c}_{left} &= \mathcal{E}_{BCT}(\{f_{left}^k\}_{k=K}^1) \\ 
\left ( \{\overleftarrow{f_{mid}^k}\}_{k=K}^1, \{\overleftarrow{f_{left}^k}\}_{k=K}^1\ \right )  &= \mathcal{D}_{BCT}(\{f_{right}^k\}_{k=K}^1, \tilde{c}_{left}) %
\end{aligned}
\end{equation}
in which now $k$ is decreasing from $K$ to $1$ and the superscript $\leftarrow$ specifically indicates the right-to-left direction.

Finally, via concatenating $\overleftarrow{f_{mid}} = \{\overleftarrow{f_{mid}^k}\}_{k=1}^K$ and $\overrightarrow{f_{mid}} = \{\overrightarrow{f_{mid}^k}\}_{k=1}^K$ along the channel dimension followed by a $1\times1$ convolutional layer,
we obtain the feature $\tilde{f}_{mid}$ related to the intermediate region. Afterwards, the horizontal concatenation over $\{\tilde{f}_{left}, \tilde{f}_{mid}, \tilde{f}_{right}\}$ becomes the input $\tilde{f}$ for the image context decoder $\mathcal{D}$. It is particularly worth noting that both the weights of the LSTM encoder $\mathcal{E}_{BCT}$ and the conditional LSTM decoder $\mathcal{D}_{BCT}$ are shared in our implementation regardless of the directions.

\vspace{-3mm}
\subsubsection{Contextual Attention on Skip Connection}
\label{attention}
\vspace{-1mm}
Even though our designs from Section~\ref{encoder_decoder} and Section~\ref{BCT} are sufficient to produce preliminary results of blending with continuous structure and coherent spatial configuration, the generated intermediate region $\tilde{I}_{mid}$ might still seem blurry or lack the texture and details. In the hope of enriching our result with the texture and details obtained from input photos, we 
adopt the contextual attention mechanism~\cite{yu2018generative} into our proposed method.

The contextual attention is originally proposed in~\cite{yu2018generative} to address image inpainting. Basically, under the image inpainting scenario (i.e.\ filling the missing region in a given image based on the information from surrounding regions that are not missing) and assuming that now the missing region has its preliminary inpainting result, the contextual attention mechanism works as follows: first, the matching scores between the patches extracted from the surrounding regions and the missing region are computed by cosine similarity, where Softmax is applied on these matching scores to get the attention scores for each patch in the missing region. Then, the patches in the missing region can be represented by the linear combination of the patches from the surrounding regions, with using the attention scores as the weights for combination. As now the missing region borrows the rich information from the whole surrounding regions, %
better inpainting results can be achieved.

When it comes to our task of wide-range image blending, we extend the contextual attention mechanism to work with the skip connection across the layers of image context encoder and decoder, by using the following analogy with respect to the original inpainting problem: we treat the feature maps of $I_{left}$ and $I_{right}$ extracted from a certain layer $L$ in the encoder as the surrounding regions, and the feature map of $\tilde{I}_{mid}$ obtained from the corresponding layer of $L$ in the decoder as the missing region (shown in Figure~\ref{fig:model}(a)). Based on such contextual attention mechanism, the feature map related to $\tilde{I}_{mid}$ in our decoder is largely enhanced by the rich information of real texture/details from $I_{left}$ and $I_{right}$, thus leading to more appealing results of blending.

\subsection{Two-Stage Training}
We provide an overview of our training procedure before stepping into the details of our loss functions. Our model learning is composed of two stages: 
\begin{enumerate}[label=(\Roman*)]
\item \textbf{Self-Reconstruction Stage:} We adopt the objective of self-reconstruction, where the two input photos $\{I_{left}, I_{right}\}$ and the intermediate region are obtained from the same image. This is achieved by first splitting a wide image vertically and equally into three parts, then taking the leftmost one-third and the rightmost one-third as $I_{left}$ and $I_{right}$ respectively, while the middle one-third can be treated as the ground truth $I_{mid}$ for the generated intermediate region $\tilde{I}_{mid}$.
\vspace{-1mm}
\item \textbf{Fine-Tuning Stage:} We keep using the objective of self-reconstruction as the previous training stage, but additionally consider another objective which is based on the training samples of having $I_{left}$ and $I_{right}$ obtained from different images (i.e.\ different scenes). As there is no ground truth of $\tilde{I}_{mid}$ now for such training samples, this additional training objective is then based on the adversarial learning. 
\end{enumerate}
\vspace{-1mm}
\noindent The rationale behind our employing two-stage training strategy is that our model can first learn to generate high quality images through the fully-guided supervised learning upon self-reconstruction in the first stage, and then focus on enhancing the ability of blending distinct images during the second stage of fine-tuning.

\subsection{Training Objectives}
\noindent\textbf{Pixel Reconstruction Loss.}
As the output panorama $\tilde{I}$ of our proposed method is composed of $\{\tilde{I}_{left}, \tilde{I}_{mid}, \tilde{I}_{right}\}$, ideally $\tilde{I}_{left}$ and $\tilde{I}_{right}$ should be identical to the input $I_{left}$ and $I_{right}$ respectively, we can therefore define the pixel reconstruction between them. Moreover, during the self-reconstruction stage where we obtain $\{I_{left}, I_{mid}, I_{right}\}$ all from the same image, naively we could also use the pixel reconstruction to enforce $\tilde{I}_{mid}$ to be the same as $I_{mid}$. However, we relax such strong constraint by applying a weighted mask $M$ when computing the pixel errors on $\tilde{I}_{mid}$, such that the pixels which are further away from the borders between $\tilde{I}_{mid}$ and $\{\tilde{I}_{left}, \tilde{I}_{right}\}$ are penalized less in order to provide greater flexibility for the image content of the intermediate region. The mask $M$ is defined as:
\begin{equation}
M(d) = \exp(-\frac{1}{2}(\frac{d}{\sigma})^2) + \exp(-\frac{1}{2}(\frac{d-d_{total}}{\sigma})^2),
\end{equation}
where $d_{total}$ is the width of $\tilde{I}_{mid}$, $\sigma=\frac{d_{total}}{4}$, and $d$ is the horizontal position of the pixel in $\tilde{I}_{mid}$ (i.e.\ the range of $d$ is $0$ to $d_{total}$ from the leftmost pixel to the rightmost one).
The pixel reconstruction loss based on the self-reconstruction, $\mathcal{L}_{pixel}^{\text{SR}}$, is then defined as: 
\begin{equation}
\begin{aligned}
\mathcal{L}_{pixel}^{\text{SR}} = &\sum
\lVert\tilde{I}_{left} - I_{left}\rVert_2 + 
\lVert\tilde{I}_{right} - I_{right}\rVert_2 \\
&+\lVert M\odot(\tilde{I}_{mid} - I_{mid})\rVert_2,
\end{aligned}
\end{equation}
where $\odot$ stands for the pixel-wise multiplication. While the additional pixel reconstruction loss used to fine-tune our model in the fine-tuning stage, $\mathcal{L}_{pixel}^{\text{FT}}$, only has the terms related to $\tilde{I}_{left}$ and $\tilde{I}_{right}$:
\begin{equation}
\mathcal{L}_{pixel}^{\text{FT}} = \sum
\lVert\tilde{I}_{left} - I_{left}\rVert_2 + \lVert\tilde{I}_{right} - I_{right}\rVert_2
\end{equation}
Please note that the summation $\sum$ used in this paper is performed over all the training data, unless otherwise specified.

\vspace{-2mm}
\paragraph{Feature Reconstruction Loss.}
As the ground truth of the intermediate region is available when performing self-reconstruction,
we can extract the feature map of the ground truth $I_{mid}$ via our image encoder $\mathcal{E}$, and encourage our estimated $\tilde{I}_{mid}$ to be identical to it. We thus define the feature reconstruction loss $\mathcal{L}_{feat\_rec}^{\text{SR}}$ as:
\begin{equation}
\mathcal{L}_{feat\_rec}^{\text{SR}} = \sum
\lVert\tilde{f}_{mid} - \mathcal{E}(I_{mid})\rVert_2.
\end{equation}

\noindent \textbf{Texture Consistency Loss.}
We adopt the regularization of implicit diversified Markov random fields (IDMRF), as used in prior works of inpainting, outpainting, and image transformation~\cite{mechrez2018contextual, wang2018image, wang2019wide}, to minimize the difference from each pixel of $F(\tilde{I}_{mid})$ to its nearest-neighbor from $F(I_{mid})$, where $F$ is a pretrained feature extractor. In other words, IDMRF encourages similar feature distribution between $\tilde{I}_{mid}$ and $I_{mid}$, hence is capable of pushing $\tilde{I}_{mid}$ to have the rich texture as shown in its ground truth $I_{mid}$. 
Please note again that this objective needs the ground truth $I_{mid}$ therefore being only included for the training examples of self-reconstruction.
The texture consistency loss $\mathcal{L}_{mrf}^{\text{SR}}$ is written as:
\begin{equation}
\mathcal{L}_{mrf}^{\text{SR}} = \text{IDMRF}(\tilde{I}_{mid}, I_{mid}),
\end{equation}
where our computation of IDMRF is identical to the one in~\cite{wang2018image}, and the pretrained features that we use for IDMRF are from the layers \texttt{relu3\_2} and \texttt{relu4\_2} of a pretrained VGG19~\cite{simonyan2014very} network.

\vspace{-2mm}
\paragraph{Feature Consistency Loss.}
As shown in Section~\ref{BCT}, our Bidirectional Content Transfer (BCT) module predicts $\{\overrightarrow{f_{mid}}, \overrightarrow{f_{right}}\}$ from $\tilde{f}_{left}$ along the direction towards $\tilde{f}_{right}$ with being conditioned on $\tilde{c}_{right}$, as well as $\{\overleftarrow{f_{left}}, \overleftarrow{f_{mid}}\}$ from $\tilde{f}_{right}$ along the opposite direction with being conditioned on $\tilde{c}_{left}$, where we denote $\overleftarrow{f_{left}} = \{\overleftarrow{f_{left}^k}\}_{k=1}^K$ and $\overrightarrow{f_{right}} = \{\overrightarrow{f_{right}^k}\}_{k=1}^K$. Although being predicted from opposite directions, the feature maps $\{\tilde{f}_{left}, \overrightarrow{f_{mid}},  \overrightarrow{f_{right}}\}$ and $\{\overleftarrow{f_{left}}, \overleftarrow{f_{mid}}, \tilde{f}_{right}\}$ ideally should be consistent to each other respectively.
We therefore introduce the feature consistency loss $\mathcal{L}_{feat\_con}$ to impose such consistency on the predictions produced by BCT module, and it is defined as:
\begin{equation}
\begin{aligned}
\mathcal{L}_{feat\_con} = \sum&
\lVert\tilde{f}_{left} - \overleftarrow{f_{left}}\rVert_2 + 
\lVert\overrightarrow{f_{mid}} - \overleftarrow{f_{mid}}\rVert_2 \\
&+
\lVert\overrightarrow{f_{right}} - \tilde{f}_{right}\rVert_2.
\end{aligned}
\end{equation}
Please note that the loss function $\mathcal{L}_{feat\_con}$ can be used for all training examples (i.e.\ no matter $I_{left}$ and $I_{right}$ are obtained from the same image or not).

\vspace{-3mm}
\paragraph{Adversarial Loss.}
Finally, we adopt the adversarial learning technique~\cite{goodfellow2014generative} to improve the realness of the generated panorama produced by our proposed method. We use the Relativistic Average Least-Square GAN (RaLSGAN~\cite{jolicoeur2018relativistic, liu2019coherent, mao2017least}) to develop our discriminator for performing adversarial learning due to its advantages of having more stable training and generating results of higher image quality. The adversarial losses for training the discriminator and the generator (i.e our full model for producing wide-range image blending) are respectively defined as:
\begin{equation}
\label{equ:dis_global_adv}
\begin{aligned}
\mathcal{L}_{adv\_D} &= \sum[\mathcal{D}_{gan}^{Ra}(I_r, \tilde{I}) - 1]^2 + [\mathcal{D}_{gan}^{Ra}(\tilde{I}, I_r) + 1]^2,\\
\mathcal{L}_{adv\_G} &= \sum\mathcal{D}_{gan}^{Ra}(\tilde{I}, I_r)^2,
\end{aligned}
\end{equation}
where $\tilde{I}$ is our model output, $I_r$ is a real image randomly chosen from the dataset, and $\mathcal{D}_{gan}^{Ra}$ is the relativistic average discriminator evolved from the typical GAN discriminator $\mathcal{D}_{gan}$:
 \begin{equation}
\label{equ:global_ra_dis}
\mathcal{D}_{gan}^{Ra}(x,y) = \mathcal{D}_{gan}(x) - \mathbb{E}_{y}\mathcal{D}_{gan}(y).
\end{equation}

\noindent\textbf{Overall Objectives.}
In summary, all the aforementioned objectives are used for the training examples of self-reconstruction (i.e.\ $\{I_{left}, I_{mid}, I_{right}\}$ are obtained from the same image) in both stages of our training procedure. While in the fine-tuning stage, for those training examples of having $I_{left}$ and $I_{right}$ obtained from different images, only part of the pixel reconstruction loss (i.e.\ $\mathcal{L}_{pixel}^{\text{FT}}$), the feature consistency loss (i.e.\ $\mathcal{L}_{feat\_con}$), and the adversarial losses (i.e.\ $\mathcal{L}_{adv\_D}$ and $\mathcal{L}_{adv\_G}$) are adopted.
Moreover, we introduce the hyperparameters $\lambda$ to weight the loss functions for controlling their balance, where we provide the detailed settings in supplementary.
Source code is available at our project page: \url{https://github.com/julia0607/Wide-Range-Image-Blending}.

%% file: 4_experiment.tex
\begin{figure*}[ht]
\centering
\footnotesize
\stackunder[5pt]{\includegraphics[width=0.7\columnwidth]{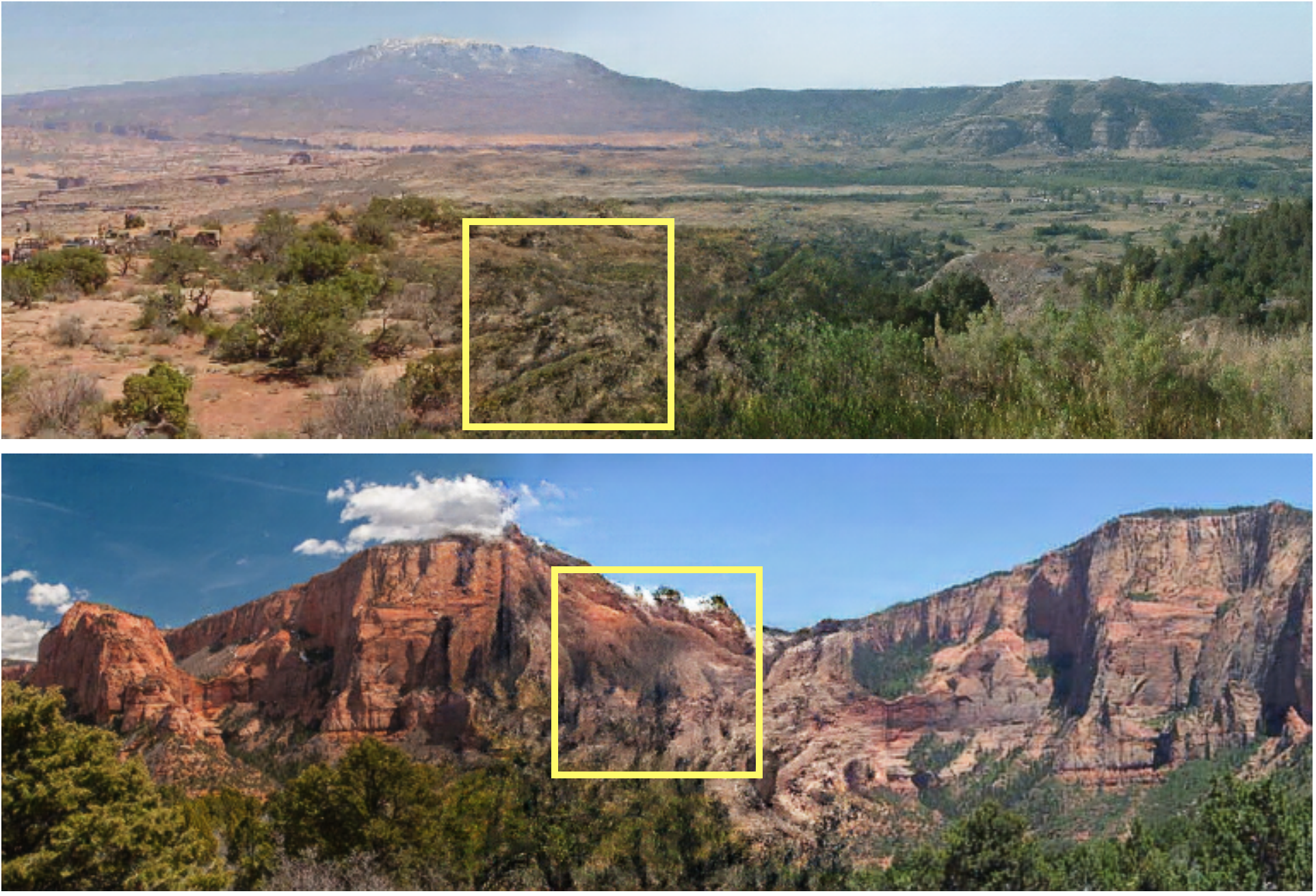}}{w/o Attention}
\hspace{-1mm}
\stackunder[5pt]{\includegraphics[width=0.1823\columnwidth]{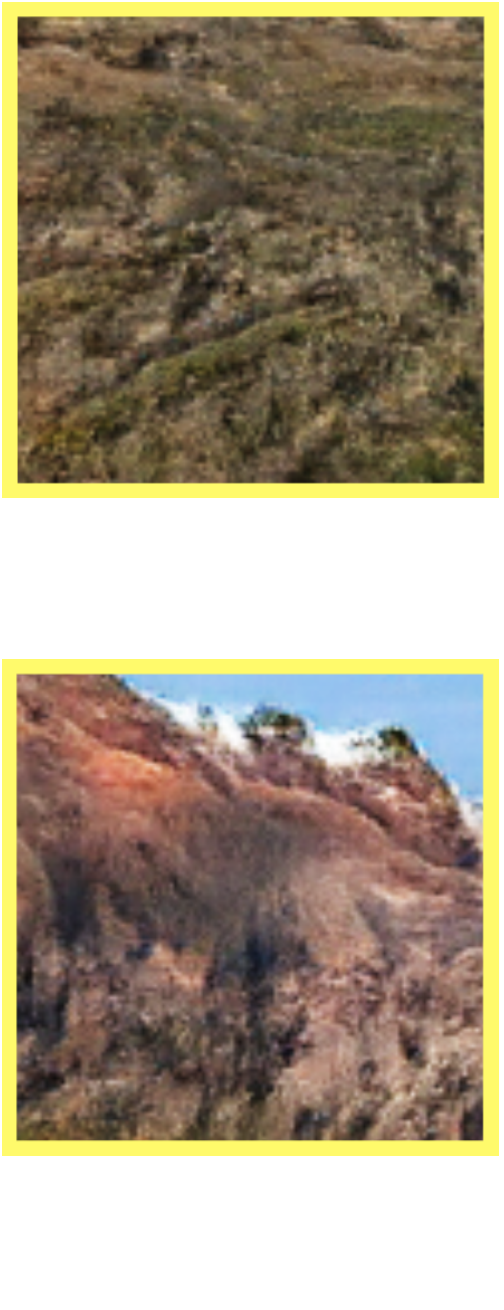}}{}
\hspace{1.5mm}
\stackunder[5pt]{\includegraphics[width=0.7\columnwidth]{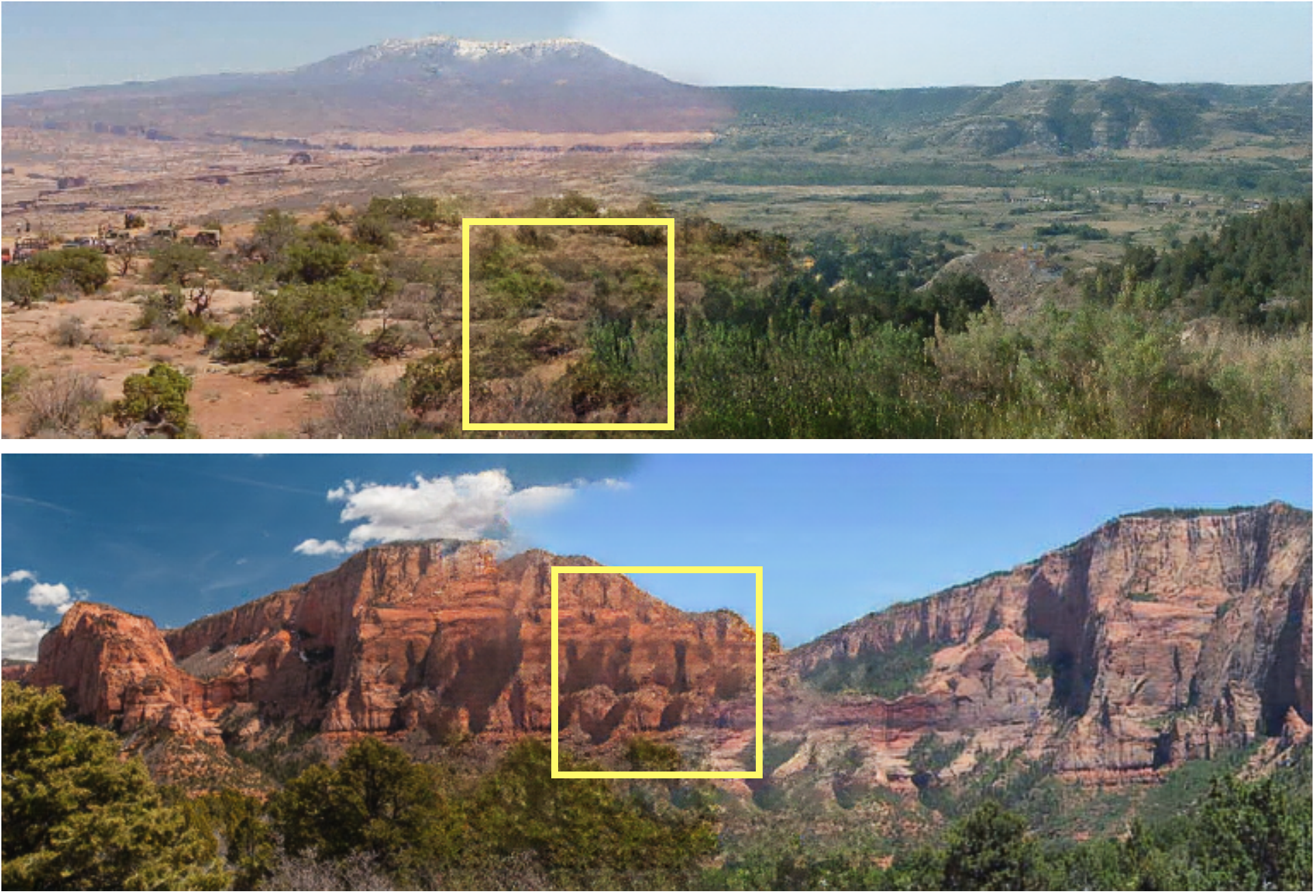}}{Full Model}
\hspace{-1mm}
\stackunder[5pt]{\includegraphics[width=0.1823\columnwidth]{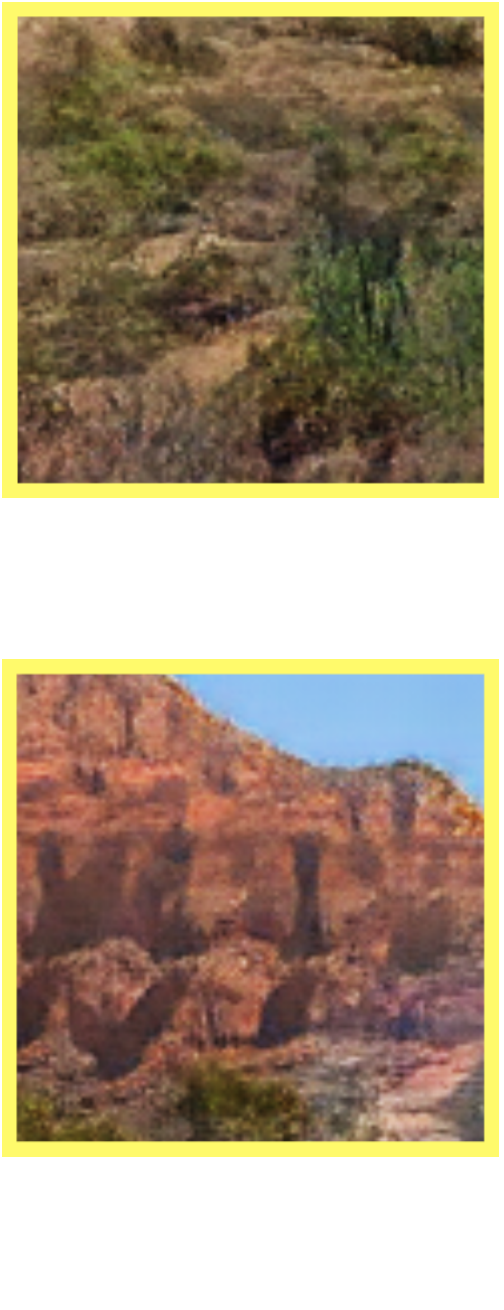}}{}
\caption{Qualitative examples for ablation study on the contextual attention mechanism  (cf. Section~\ref{ablation_study}).}
\label{fig:ablation}
\end{figure*}

\begin{table}[t]
\centering
\setlength{\tabcolsep}{1.6mm}{
\begin{tabular}{c|c|ccc}
\multicolumn{2}{c|}{\multirow{2}{*}{Model Variants}} & \multirow{2}{*}{FID(↓)} & \multicolumn{2}{c}{KID(↓)}        \\
\multicolumn{2}{c|}{}                       &          & mean   & std    \\ \hline\hline
\multicolumn{2}{c|}{w/o Attention}                    & 35.86     & 0.0105    & 0.0005  \\ \cline{1-5}
\multicolumn{2}{c|}{w/o $\mathcal{L}_{pixel}$}        & 42.14     & 0.0148    & 0.0008  \\ \cline{1-5}
\multicolumn{2}{c|}{w/o $\mathcal{L}_{feat\_rec}$}    & 37.21     & 0.0124    & 0.0006  \\ \cline{1-5}
\multicolumn{2}{c|}{w/o $\mathcal{L}_{mrf}$}          & 44.74     & 0.0216    & 0.0007  \\ \cline{1-5}
\multicolumn{2}{c|}{w/o $\mathcal{L}_{feat\_con}$}    & 45.29     & 0.0224    & 0.0009  \\ \cline{1-5}
\multicolumn{2}{c|}{w/o $\{\mathcal{L}_{adv\_D}, \mathcal{L}_{adv\_G}\}$}              & 57.62     & 0.0382    & 0.0012 \\ \hline \hline
\multirow{2}{*}{Full Model}       & SR Stage             & 46.30  & 0.0218 & 0.0010 \\ \cline{2-5}
                                  & FT Stage (Final)     & 36.13 & 0.0116 & 0.0005
\end{tabular}
}
\caption{Ablation study on each of our model designs and our two-stage training procedure.}
\label{tab:ablation_study_table}
\end{table}

\begin{table}[ht]
\setlength{\tabcolsep}{1.2mm}{
\begin{tabular}{c|c|ccc}
\multicolumn{2}{c|}{\multirow{2}{*}{Method}} & \multirow{2}{*}{FID(↓)} & \multicolumn{2}{c}{KID(↓)}        \\
\multicolumn{2}{c|}{}                       &          & mean   & std    \\ \hline\hline
\multirow{5}{*}{Inpainting}  & CA~\cite{yu2018generative}               & 91.87  & 0.0745 & 0.0022 \\ \cline{2-5} 
                             & PEN-Net~\cite{zeng2019learning}          & 159.70 & 0.1151 & 0.0020 \\ \cline{2-5} 
                             & StructureFlow~\cite{ren2019structureflow}& 138.13 & 0.2168 & 0.0023 \\ \cline{2-5}
                             & HiFill~\cite{yi2020contextual}           & 139.39 & 0.1230 & 0.0028 \\ \cline{2-5}
                             & ProFill~\cite{zeng2020high}              & 46.53  & 0.0326 & 0.0011 \\ \hline
\multirow{2}{*}{Outpainting} & SRN~\cite{wang2019wide}                  & 70.94  & 0.0392 & 0.0012 \\ \cline{2-5} 
                             & Yang \etal~\cite{yang2019very}           & 82.69  & 0.0446 & 0.0012 \\ \hline
\multicolumn{2}{c|}{Ours}                    & \textbf{36.13}        & \textbf{0.0116} & \textbf{0.0005}
\end{tabular}
}
\caption{Quantitative comparison with respect to various baselines from image impainting and outpainting.}
\label{table:quantitative}
\end{table}

\begin{figure*}[ht]
\centering
\footnotesize
\stackunder[5pt]{\includegraphics[width=0.29\columnwidth]{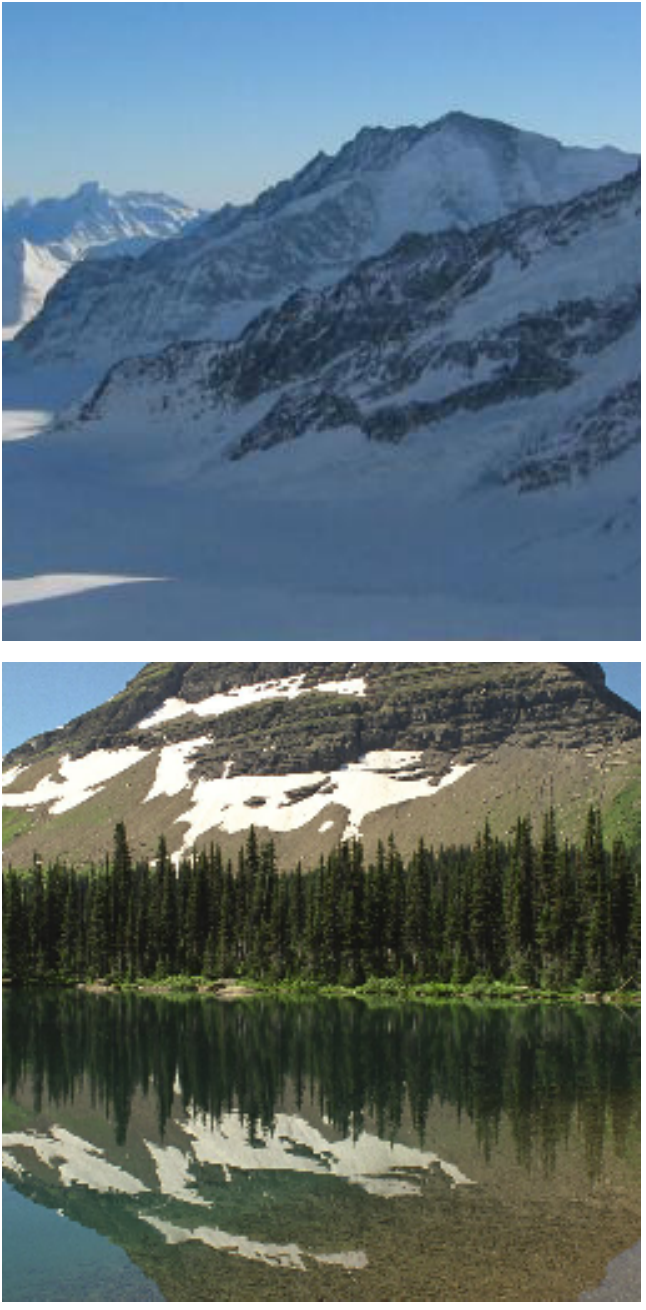}}{First Input Photo}
\hspace{-1mm}
\stackunder[5pt]{\includegraphics[width=0.29\columnwidth]{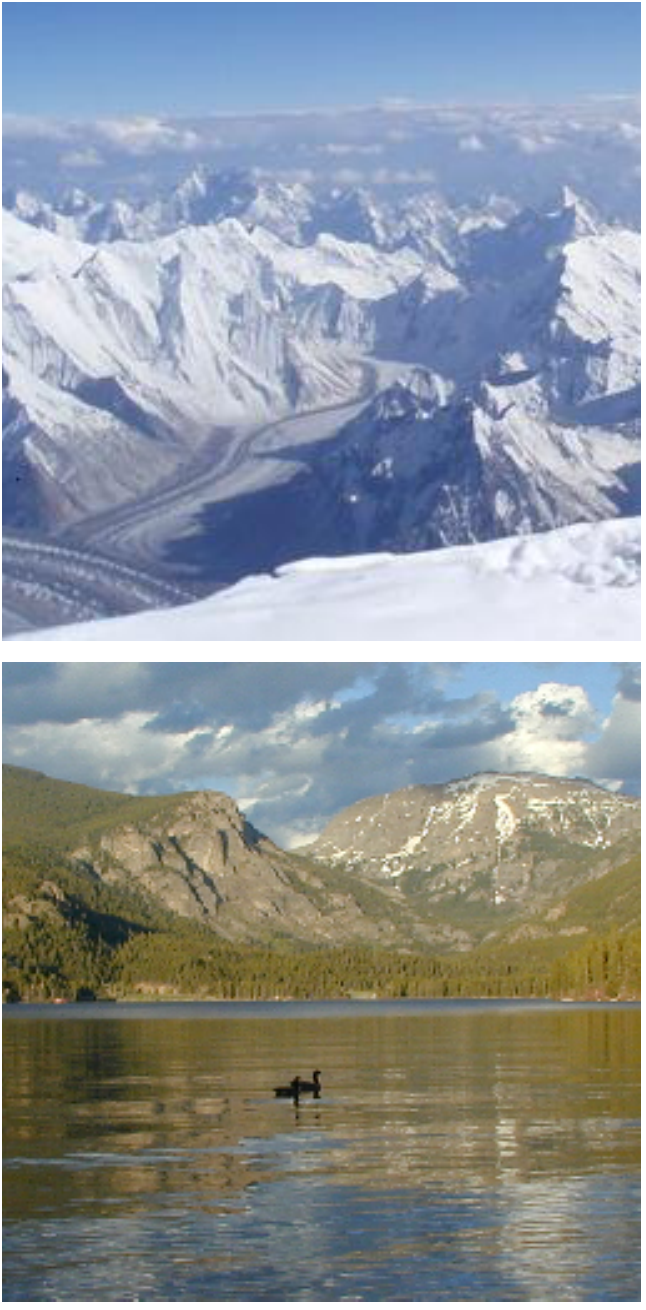}}{Second Input Photo}
\stackunder[5pt]{\includegraphics[width=1.45\columnwidth]{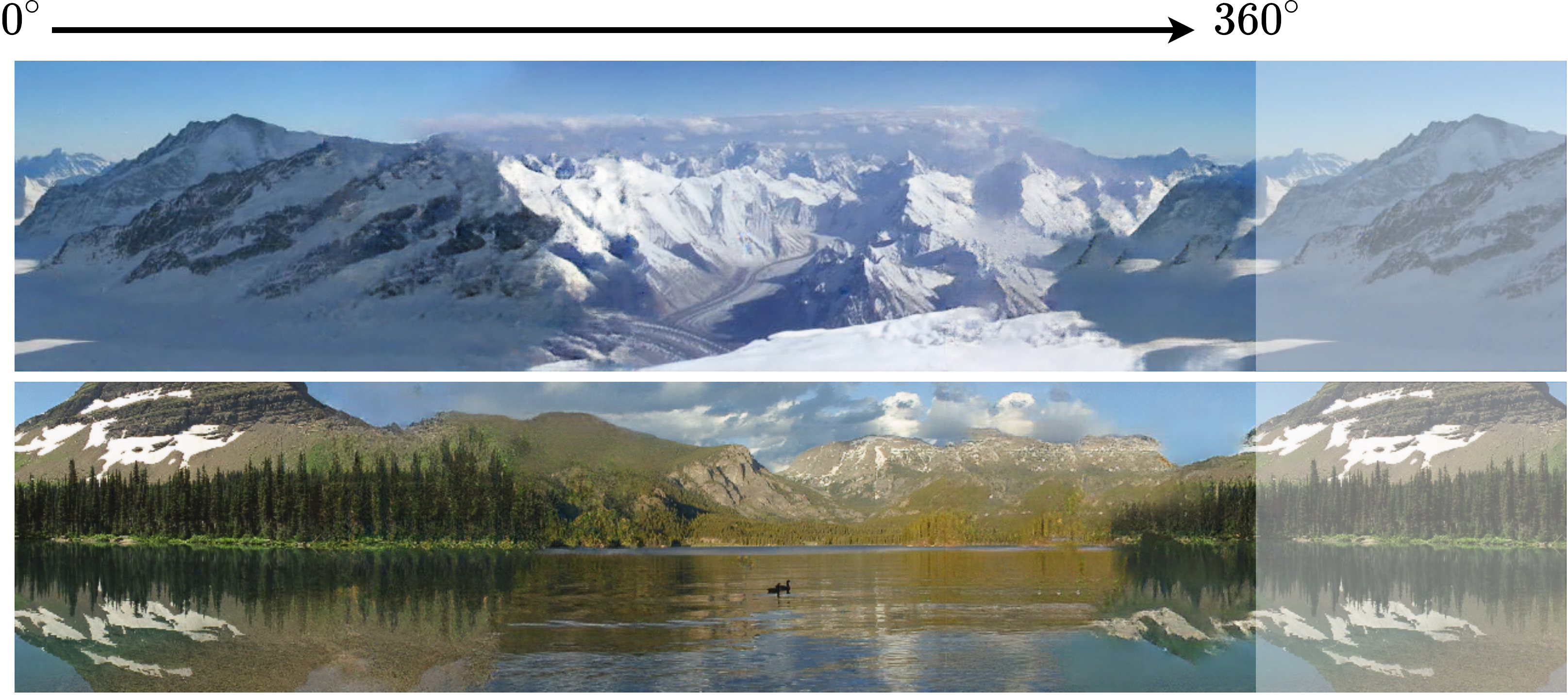}}{Panorama Result}
\caption{Given two different input images (the first two columns), our method can construct a full panoramic image (the third column) that provides cyclic view by stitching the two blending results generated from two opposite spatial arrangements.}
\label{fig:panorama}
\end{figure*}

\begin{figure*}[t]
	\centering
	\includegraphics[width=0.96\textwidth]{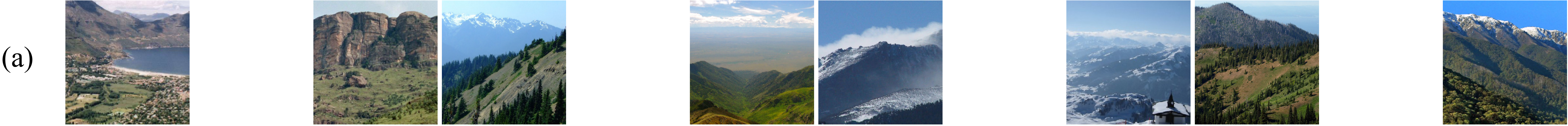}
    \includegraphics[width=0.96\textwidth]{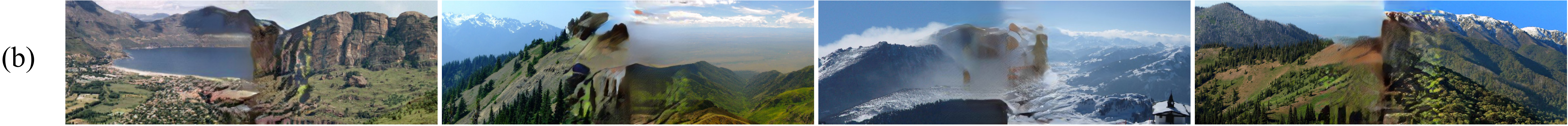}
    \includegraphics[width=0.96\textwidth]{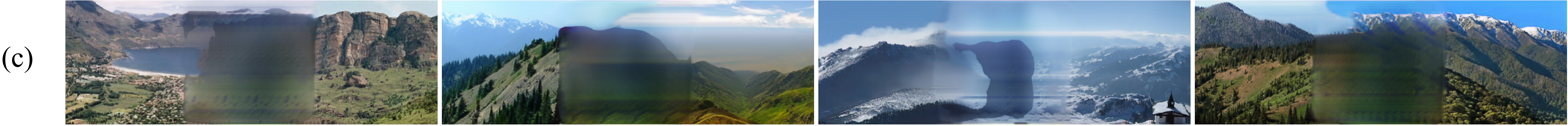}
    \includegraphics[width=0.96\textwidth]{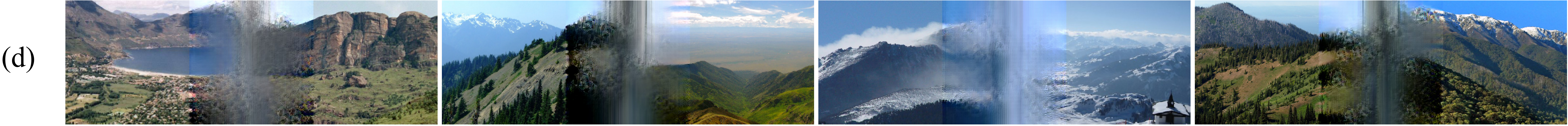}
    \includegraphics[width=0.96\textwidth]{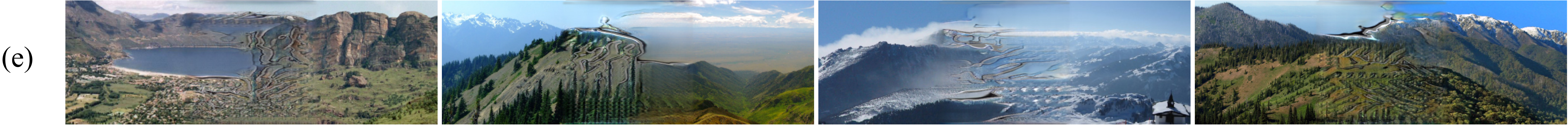}
    \includegraphics[width=0.96\textwidth]{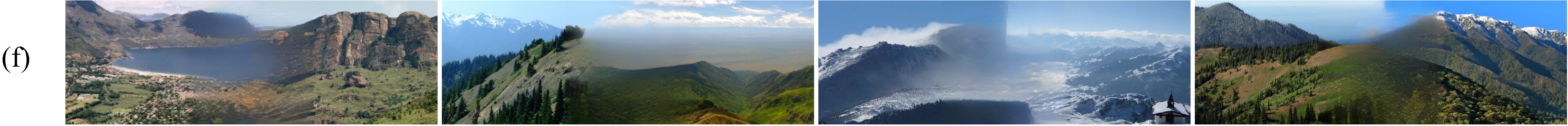}
    \includegraphics[width=0.96\textwidth]{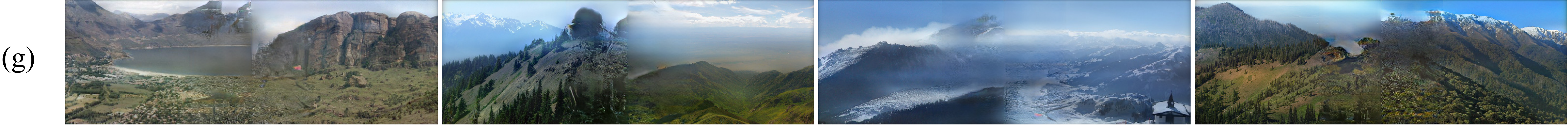}
    \includegraphics[width=0.96\textwidth]{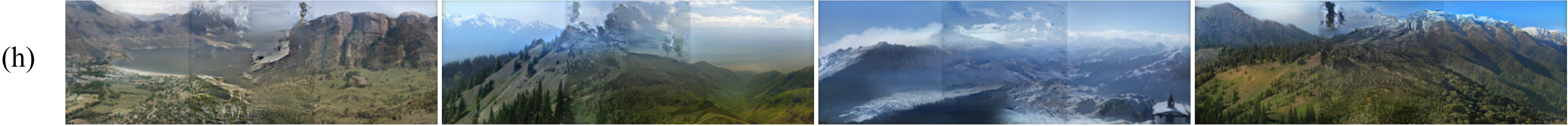}
    \includegraphics[width=0.96\textwidth]{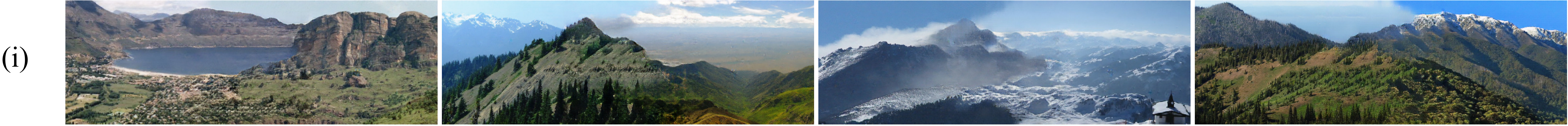}
    \caption{Qualitative comparison with baselines of image inpainting and image outpainting: (a) input images, (b) CA~\cite{yu2018generative}, (c) PEN-Net~\cite{zeng2019learning}, (d) StructureFlow~\cite{ren2019structureflow}, (e) HiFill~\cite{yi2020contextual}, (f) ProFill~\cite{zeng2020high}, (g) SRN~\cite{wang2019wide}, (h) Yang \etal~\cite{yang2019very}, and (i) Ours.}
	\label{fig:qualitative}
\end{figure*}

\section{Experiments}
\label{sec:experiments}

\noindent\textbf{Dataset.}
We adopt the scenery dataset proposed by~\cite{yang2019very} for conducting our experiments, in which this dataset consists of 5040 training images and 1000 testing images.
For building up our training samples for the use of self-reconstruction objective, we randomly crop the training image into the size of $256\times768$, where its leftmost and rightmost $256\times256$ regions serve as the two input photos (i.e.\ $I_{left}$ and $I_{right}$) and the middle region is the ground truth for $\tilde{I}_{mid}$.
While for the additional training samples used in the fine-tuning stage (i.e.\ $I_{left}$ and $I_{right}$ obtained from different images), we would like the model learning to blend the images from different scenes. However, if the two input photos are overly distinct from each other, the learning would become too difficult to achieve. We therefore propose the following manner to prepare the training samples: We first crop many $256\times256$ regions from the training images. Then for each cropped region, we adopt the Learned Perceptual Image Patch Similarity (LPIPS) metric~\cite{zhang2018unreasonable} to find the first three of its most similar cropped regions from other images, thus forming the input pairs for our model training. 

\vspace{-5mm}
\paragraph{Metrics.}
We use Fr\'{e}chet Inception distance (FID~\cite{heusel2017gans}) and Kernel Inception Distance (KID~\cite{binkowski2018demystifying}) as metrics for our quantitative evaluation. FID is commonly used to measure the fidelity and diversity of generative images with respect to the real ones, where their inception features are fitted by Gaussian and the Fr\'{e}chet distance are computed between Gaussians; KID is similar to FID but instead uses the squared Maximum-Mean-Discrepancy between features. Both FID and KID are the lower the better. Please note that, with considering the fact that a huge portion in our output panorama requires only reconstruction of the input photos, which is not the main target of our evaluation, we therefore crop the central area of size $256\times512$ from the output panorama of size $256\times768$ for performing evaluation and comparison (as well for the baselines).

\subsection{Ablation Study}
\label{ablation_study}
To better understand the contribution of each component as well as each objective function in our proposed model, we conduct ablation study by using different model variants, where they are trained with the same training strategy (i.e.\ our two-stage training procedure), experimental settings, and the dataset, but having a specific component/objective removed.
The quantitative results are provided in Table~\ref{tab:ablation_study_table}, we can see that except for the contextual attention mechanism, the variants of removing each design or objective from our full model result in worse performance, proving that these designs/objectives are able to boost our model learning and improve the quality of our generated panoramas. 
We further study the contextual attention mechanism from qualitative results, with some examples shown in Figure~\ref{fig:ablation}. Although removing the attention mechanism seems to create gradient effects and provide smoother transition for blending, the image quality of the generated intermediate region is unsatisfactory. 
On the contrary, our full model with introducing the contextual attention mechanism generates more delicate blending results with rich texture and exquisite details.
Besides conducting ablation study on our model designs, we provide quantitative results of our full model with different training strategies in the last two rows of Table~\ref{tab:ablation_study_table} as well. The model variant of excluding the fine-tuning stage has inferior performance with respect to the full model with having the complete two-stage training, thus verifying the effectiveness of the our proposed two-stage procedure for model training.

\subsection{Quantitative Results}
\label{quantitative_results}
We make the quantitative comparison with respect to several state-of-the-art models for image inpainting and outpainting. 
Basically, we directly adopt inpainting models to the wide-range image blending task by treating the two input photos as the given context and the intermediate region as the missing area to be filled. As for the adopting outpaining models, we first apply them on the two input photos individually for generating new contents beyond the boundaries towards the intermediate region, then we employ an state-of-the-art image blending method (i.e.\ GP-GAN~\cite{wu2019gp}) to blend the two extrapolated images.
The results of quantitative comparison are shown in Table~\ref{table:quantitative}, where our proposed method clearly outperforms the baselines, showing that the proposed task of wide-range image blending is worth-discussing and difficult for the existing approaches, and that our proposed method is capable of solving the task. 

\subsection{Qualitative Results}
\label{qualitative_results}
Figure~\ref{fig:qualitative} shows some examples of blending results obtained from the baselines (e.g.\ the image inpainting methods such as CA~\cite{yu2018generative}, PEN-Net~\cite{zeng2019learning}, StructureFlow~\cite{ren2019structureflow}, HiFill~\cite{yi2020contextual}, and ProFill~\cite{zeng2020high}, as well as the outpainting ones from SRN~\cite{wang2019wide} and Yang \etal~\cite{yang2019very}) and our proposed model. Image inpainting methods create significant artifacts, distorted structures, and blurry textures since they are unable to deal with the large missing regions, and to cooperate the image context coming from different input photos either. Similarly, image outpainting methods are deficient in collaborating the distinct contents from the two input photos even though they can extrapolate the input images with better quality in comparison to inpainting methods. Both inpainting and outpainting baselines create discontinuous structure in the intermediate region and fail to smoothly merge the two input photos into a realistic panorama. On the contrary, our blending results demonstrate that our proposed method is able to generate a high-quality and realistic intermediate region, which merges the two input photos with seamless transition while retaining a reasonable semantic configuration. Furthermore, in Figure~\ref{fig:panorama} we provide some examples of an interesting application of our proposed method, which utilizes two input photos to generate a full panoramic image that provides a complete cyclic view. Our resultant full panorama not only shows smooth transitions between the input photos but also displays realistic details. 
Please refer to supplementary materials for more results.

%% file: 5_conclusion.tex
\section{Conclusion}
\label{sec:conclusion}
In this paper, we propose a new research problem in image processing, \emph{Wide-Range Image Blending}, as well as an effective model with several novel designs to adequately deal with such new problem. 
We provide experimental evidence to prove that directly applying existing methods of related topics (such as image inpainting or outpainting) leads to poor results, while our proposed method is able to generate novel image content for smoothly merging two different images into a favorable panoramic image.\\

\noindent\textbf{Acknowledgement.} This project is supported by MediaTek Inc., MOST 110-2636-E-009-001, MOST 110-2634-F-009-018, and MOST-110-2634-F-009-023. We are grateful to the National Center for High-performance Computing
for computer time and facilities.